\documentclass{article}

\usepackage[accepted]{icml2026}\usepackage{microtype}
\usepackage{graphicx}
\usepackage{booktabs}
\usepackage{amsmath}
\usepackage{amssymb}
\usepackage{hyperref}
\usepackage{url}
\usepackage{xcolor}
\usepackage{xspace}
\usepackage{enumitem}
\usepackage{multirow}
\usepackage{tabularx}
\usepackage{placeins}

\definecolor{darkblue}{rgb}{0, 0, 0.5}
\hypersetup{colorlinks=true, citecolor=darkblue, linkcolor=darkblue, urlcolor=darkblue}

\newcommand{\method}{Parallel WebBench\xspace}

\icmltitlerunning{When Web Agents Finish but Still Fail}

\begin{document}

\twocolumn[
\icmltitle{When Web Agents Finish but Still Fail: Reproducible Triggers and Trace Diagnostics for Parallel Web Exploration}

\begin{icmlauthorlist}
\icmlauthor{Aagam Sogani}{uw}
\icmlauthor{Botao Rui}{uw}
\icmlauthor{Swetha Vaidyanathan}{uw}
\icmlauthor{Rishi Agarwal}{uw}
\icmlauthor{Minghao Yan}{uw}
\icmlauthor{Shivaram Venkataraman}{uw}
\end{icmlauthorlist}

\icmlaffiliation{uw}{University of Wisconsin--Madison, Madison, WI, USA}

\icmlcorrespondingauthor{Aagam Sogani}{sogani.aagam@gmail.com}

\vskip 0.3in
]

\printAffiliationsAndNotice{}

\begin{abstract}
Long-horizon web agents often fail in ways hidden by final-answer evaluation: they may visit useful pages, produce a well-formed answer, and terminate confidently while still missing fields, over-including unsupported items, or relying on stale evidence. We study these failures with \method, a parallel web-exploration benchmark containing 1,679 verified records: 350 manually curated parallel tasks and 1,329 reconstructed records with verified URL-based trajectories. We train WebExplorer-style agents with GRPO under human-only, balanced human-synthetic, and synthetic-heavy data mixtures. At 16k context and 16 interaction rounds, the best GRPO model improves completion over WebExplorer-8B from 50.7\% to 96.0\% and GPT-4.1-mini-judged element-wise F1 from 0.2489 to 0.4529, but binary accuracy remains far below completion. Trace-level analysis identifies three persistent failure modes: context-bound search loops, premature termination on partial answers, and synthesis collapse after relevant evidence has already been retrieved. These results show that synthetic-data GRPO reduces abstention and improves partial correctness, but leaves a completion-correctness gap that requires evidence-grounded coverage and synthesis diagnostics.
\end{abstract}

\section{Introduction}
\label{sec:intro}

Web agents are increasingly evaluated on tasks that require navigation, information gathering, and long-horizon reasoning over real websites \citep{nakano2021webgpt,yao2023react,zhou2023webarena,he2024webvoyager}. These agents rarely fail in a single obvious step. A bad search query can lead to a misleading page, a partially retrieved fact can be mistaken for a complete answer, and an early assumption can persist through many later tool calls. By the time the agent emits a final response, the reason for the failure may no longer be visible from the answer alone.

Many existing web-agent and web-exploration benchmarks focus on sequential information-seeking or task-completion \citep{yao2022webshop,deng2023mind2web,zhou2023webarena,koh2024visualwebarena,drouin2024workarena,lu2024weblinx,he2024webvoyager,wu2025webwalker,wei2025browsecomp}; an agent follows a single navigation path to recover a single target answer, similar in spirit to multi-hop and knowledge-intensive QA settings where systems retrieve evidence for a target response \citep{yang2018hotpotqa,petroni2021kilt,gao2023enabling}. In contrast, many realistic web tasks naturally decompose into multiple sub-goals. A user may ask for several events from a conference website, multiple committee members from different pages, or a list of product attributes scattered across subpages. These tasks require the agent to retrieve information from multiple sources, keep subproblems separate, and aggregate the results into a structured final answer. We call this setting \emph{parallel web exploration}.

Parallel web exploration exposes a different class of failures from standard single-answer navigation. An agent can complete the task format while still failing the task content: it may answer only the easiest subparts, over-include unsupported entities, collapse multiple rows into one repeated value, or stop searching once it has found a plausible partial answer. These failures are especially important for agentic AI because they are not captured well by completion rate alone and relate to broader concerns in attribution and fine-grained factuality evaluation \citep{rashkin2023measuring,min2023factscore,gao2023enabling}. A model that emits more final answers can look stronger while still being unreliable.

In this work, we introduce \method, a benchmark and construction pipeline for parallelizable web exploration. Starting from root URLs and task patterns inspired by prior web-exploration datasets such as WebWalkerQA, WebExplorer, and WebSailor \citep{wu2025webwalker,liu2025webexplorer,li2025websailor}, we generate multi-field questions that require collecting several related pieces of information, often from different subpages, and combining them into a structured output. We then curate these questions using a manual verification protocol that records answerability, the number of sequential steps, the maximum parallel degree, visited URLs, and golden answers.

Manual verification is expensive, so we also built an automated reconstruction pipeline to verify and repair existing web QA records. This pipeline filters English records, checks whether the answer remains valid on the live web, classifies records by path and answer validity, reconstructs broken URL paths when possible, and strictly verifies each link chain. This emphasis on live evidence is motivated by prior work on browser-assisted QA, changing world knowledge, and source attribution \citep{nakano2021webgpt,vu2024freshllms,rashkin2023measuring}. The resulting dataset contains 1,679 verified records: 350 manually curated parallel tasks and 1,329 automatically reconstructed records with verified trajectories.

We train WebExplorer-style agents \citep{liu2025webexplorer} using Group Relative Policy Optimization (GRPO) \citep{schulman2017proximal,shao2024deepseekmath} under three data mixtures: human-only supervision, balanced human-synthetic supervision, and synthetic-heavy supervision. We also study reward design for structured web-agent training, comparing exact-match reward signals with denser partial-credit reward styles, including LLM-judge-based scoring used in our reward-signal analysis. For the final evaluation, we apply the same GPT-4.1-mini LLM-as-judge protocol to all systems, including baselines and our GRPO-trained models, and report two correctness metrics: Binary Accuracy, which requires the full structured answer to be judged correct, and Element-wise F1, which gives partial credit over recovered answer elements \citep{zheng2023judging,liu2023geval}.

Our main finding is that GRPO and synthetic data substantially improve completion and partial correctness, but do not remove core agentic failure modes, echoing broader observations that outcome-level optimization and human-feedback training can improve behavior while leaving reliability gaps \citep{ouyang2022training,lightman2024let,shao2024deepseekmath}. At 16k context and 16 interaction rounds, our best GRPO model improves completion from 50.7\% to 96.0\% and Element-wise F1 from 0.2489 to 0.4529 over the base WebExplorer-8B model. Yet Binary Accuracy remains much lower than completion, and increasing the inference budget does not improve correctness monotonically. Trace diagnostics show why: trained agents often finish more confidently without fully solving the task.

We keep these reported correctness metrics separate from training-time rewards. The three main GRPO runs reported in Table~\ref{tab:baseline_all} are trained with a partial-credit reward based on fuzzy matching over dictionaries, lists, and scalar fields. The reward-signal diagnostic in Appendix~\ref{app:reward_signal} compares exact-match scoring with a WebExplorer-8B judge score on logged training samples, and is used only to illustrate reward sparsity.

Our contributions are as follows:
\begin{itemize}[nosep,leftmargin=1.5em]
    \item We define a parallel web-exploration setting that exposes a completion-correctness gap: agents emit structured final answers while missing, over-including, or misbinding retrieved facts.
    \item We introduce \method, a benchmark of 1,679 verified web-exploration records with answer, trajectory, and URL-level verification metadata.
    \item We identify three reproducible failure triggers: canonical-phrase search loops, premature termination on list-valued questions, and synthesis collapse in multi-row structured outputs.
    \item We provide trace-level diagnostics for each failure mode, including operational detectors based on redundant search behavior, round usage, element-level recall, and evidence-to-answer binding.
    \item We evaluate interventions including GRPO, synthetic supervision, context scaling, and reward-signal analysis, showing which failures improve, which persist, and what trade-offs each intervention introduces.
\end{itemize}

The central result is therefore constructive: our interventions do not solve parallel web exploration, but they reveal how agents fail and point to repair targets for future work, including coverage-aware stopping, evidence-bound synthesis, and diagnostics that separate finishing from actually solving the task \citep{gao2023enabling,rashkin2023measuring,min2023factscore}.

\section{Related Work}
\label{sec:related}

\paragraph{Web agent benchmarks.}
Prior web-agent benchmarks evaluate agents in simulated web environments, shopping tasks, realistic browser environments, enterprise software, multimodal web tasks, live websites, and general instruction-following settings \citep{yao2022webshop,deng2023mind2web,zhou2023webarena,koh2024visualwebarena,drouin2024workarena,lu2024weblinx,zheng2024gpt,he2024webvoyager,liu2024agentbench,xie2024osworld,zheng2024webolympus}. More recent benchmarks stress live or long-horizon information seeking, including web traversal from root websites and hard-to-find browsing tasks \citep{wu2025webwalker,wei2025browsecomp}. These benchmarks have been important for measuring tool use and search persistence, but most still emphasize sequential navigation or single-answer retrieval. \method targets parallel web exploration, where multiple subgoals must be retrieved and synthesized into a single structured answer.

\paragraph{Long-horizon web-agent training.}
Recent work trains open-source web agents with synthesized browsing data, reinforcement learning, and self-improvement-style agent loops. WebExplorer generates challenging query-answer pairs through model-based exploration and iterative query evolution, then trains WebExplorer-8B with supervised fine-tuning followed by GRPO \citep{liu2025webexplorer}. WebSailor similarly studies complex information-seeking tasks with high uncertainty and uses post-training to improve long-horizon web reasoning \citep{li2025websailor}. In contrast, Reflexion studies the use of verbal feedback to improve language agents \citep{shinn2023reflexion}. Our work differs by focusing less on maximizing final benchmark accuracy and more on diagnosing why trained web agents still fail after they learn to emit final answers.

\paragraph{Reasoning-action scaffolds and reward design.}
Our agent format follows the ReAct paradigm, where language models interleave reasoning traces with tool actions and observations \citep{yao2023react}. This sits within a broader line of work on tool use and browser-assisted language agents \citep{schick2023toolformer,nakano2021webgpt}. For policy optimization, we use GRPO-style training, which builds on the policy-gradient family of methods, including PPO, and has been used to improve reasoning behavior in language models \citep{schulman2017proximal,shao2024deepseekmath}. Training web agents also raises reward-design issues: exact-match rewards can be sparse for structured multi-field outputs. At the same time, human feedback and process supervision show that reward granularity affects learned behavior \citep{ouyang2022training,lightman2024let}. In contrast, dense format or process rewards can create reward-hacking pressure \citep{amodei2016concrete}. We therefore analyze not only final scores, but also trace-level failure modes and intervention trade-offs.

\paragraph{LLM-as-judge evaluation.}
LLM-as-judge evaluation is widely used to score open-ended model outputs when an exact match is too brittle \citep{zheng2023judging,liu2023geval}. We use GPT-4.1-mini as a blinded semantic judge for Binary Accuracy and element-wise F1 across all models. Because judge-based evaluation can introduce its own artifacts, including bias and rubric sensitivity \citep{wang2024large,kim2024prometheus}, we report completion separately from correctness and include a manual judge-validation audit in Appendix~\ref{app:judge_validation}.

\section{Benchmark Construction}
\label{sec:data_curation}

\method is built to evaluate parallel web exploration: tasks where the answer cannot be recovered from a single target page, but instead requires collecting multiple fields from several reachable pages and aggregating them into a structured output. This emphasis on evidence collection and structured answerability builds on multi-hop QA, knowledge-intensive benchmarks, fact verification, and citation-grounded generation \citep{yang2018hotpotqa,petroni2021kilt,thorne2018fever,gao2023enabling}. Each record contains a root URL, question, verified answer, and trajectory metadata.

\paragraph{Parallel degree as task difficulty.}
We use maximum parallel degree as a lightweight measure of task structure. A degree-$d$ task requires the agent to maintain up to $d$ distinct evidence branches before producing the final answer, a structure related to multi-hop evidence use and deliberate decomposition in reasoning systems \citep{yang2018hotpotqa,yao2023tree}. Higher-degree tasks are generally harder because the agent must preserve recall across more independent retrieval targets, avoid mixing facts between branches, and bind each retrieved fact to the correct output field during synthesis. This makes parallel degree different from trajectory length: a long single-path task can fail through search drift, while a high-degree task can fail even after relevant evidence has been found, because the final answer must correctly aggregate multiple partially independent subgoals.

\paragraph{Human-verified parallel tasks.}
We first generate candidate questions from root URLs using few-shot prompting. The prompt rewrites single-target web tasks into multi-field questions with more explicit decomposition, for example, by asking for several related entities, dates, or attributes from different subpages. Human annotators then verify whether each question is answerable on the root website and record four fields: the sequential step count, the maximum parallel degree, the visited URLs grouped by step, and a structured golden answer. If a question is partially answerable or ill-posed, annotators either repair it minimally or mark it unanswerable with a short failure description. This process yields 350 manually verified parallel tasks.

\paragraph{Automated reconstruction from prior QA data.}
Manual verification is high quality but expensive. To scale the benchmark, we reconstruct additional records from a larger pool of prior web-exploration QA data. Each source record contains a root URL, question, answer, target source URLs, and text-based navigation labels. We verify whether the answer still appears on the live web, classify records by whether their original path and answer remain valid, and reconstruct URL-based trajectories when the answer source remains valid but the original navigation path is broken. Every reconstructed trajectory is strictly checked as a link-chain: for each consecutive URL pair, the next URL must appear among the extracted links from the previous page. This verification step is aligned with evidence-centered evaluation traditions in fact verification, attribution, and knowledge-intensive QA \citep{thorne2018fever,rashkin2023measuring,petroni2021kilt,gao2023enabling}. Full annotation and reconstruction details are in Appendix~\ref{app:path_verification}.

\paragraph{Dataset yield.}
Table~\ref{tab:recon_stats} summarizes the reconstruction yield. From 13,579 raw records, language filtering retains 3,099 English records. Agentic verification classifies 612 as Category~A, where path and answer are verified; 1,289 as Category~B, where the source is verified but the original path is broken; and 1,198 as Category~C, where verification fails. Path reconstruction succeeds for 717 Category~B inputs, yielding 1,329 reconstructed records. In addition to the 350 manually verified parallel tasks, \method contains 1,679 verified records.

\begin{table}[t]
\centering
\small
\setlength{\tabcolsep}{4pt}
\caption{Reconstruction pipeline yield. Cat.~A has a verified path and answer; Cat.~B has a verified source but broken original path; Cat.~C fails verification. The final benchmark combines 1,329 reconstructed records with 350 manually verified parallel tasks.}
\label{tab:recon_stats}
\begin{tabular}{@{}lrrr@{}}
\toprule
\textbf{Stage} & \textbf{1st} & \textbf{2nd} & \textbf{Total} \\
\midrule
Raw      & 6{,}789 & 6{,}790 & 13{,}579 \\
English  & 1{,}590 & 1{,}509 & 3{,}099 \\
\midrule
Cat.~A & 371 & 241 & 612 \\
Cat.~B & 459 & 830 & 1{,}289 \\
Cat.~C & 760 & 438 & 1{,}198 \\
\midrule
Success & 250 & 467 & 717 \\
Unanswerable & 206 & 363 & 569 \\
\midrule
\textbf{Reconstructed} & \textbf{621} & \textbf{708} & \textbf{1{,}329} \\
\bottomrule
\end{tabular}
\end{table}

\section{Experiments}
\label{sec:experiments}

\subsection{Experimental Setup}
\label{sec:exp_setup}

We evaluate \method from three perspectives. First, we test whether GRPO training improves web-agent performance relative to base WebExplorer- and WebWalker-style baselines. Second, we compare training data mixtures to study how much synthetic supervision helps relative to human-verified supervision. Third, we analyze failures at the trajectory level to determine whether improvements in completion correspond to genuine task solving.

\paragraph{Training configuration.}
We train WebExplorer-style agents using Group Relative Policy Optimization (GRPO), a relative policy-optimization variant related to PPO-style policy-gradient training \citep{schulman2017proximal,shao2024deepseekmath}. Each training example consists of a root URL, a question, and a verified structured answer. During rollout, the agent interacts with web tools, gathers evidence, and emits a final structured answer. We use the same base model, scaffold, and training budget across three data mixtures, varying only the composition of human and synthetic data.

\paragraph{Training data mixtures.}
Table~\ref{tab:data_splits} summarizes the three training runs. The \textsc{Human} split uses only manually verified human-curated tasks. The \textsc{Balanced} split mixes human tasks with equal amounts of Category~A and Category~B synthetic data. The \textsc{Synth-Heavy} split further reduces the amount of human supervision and increases the proportion of synthetic data. Category~A synthetic records are easier: both the source path and answer can be verified directly. Category~B records are harder: the answer source remains valid, but the original path is broken and must be reconstructed. This split design lets us test whether synthetic supervision improves genuine web-agent behavior or mainly teaches the model to complete easier automatically generated tasks.

\begin{table}[t]
\centering
\small
\caption{GRPO training data mixtures. All runs use the same training budget; only the human/synthetic composition changes. Category~A synthetic records have verified paths and answers, while Category~B records require path reconstruction.}
\label{tab:data_splits}
\begin{tabular}{@{}llrrrr@{}}
\toprule
\textbf{Run} & \textbf{Data mixture} & \textbf{Human} & \textbf{Cat. A} & \textbf{Cat. B} & \textbf{Steps} \\
\midrule
1 & \textsc{Human}       & ${\sim}300$ & 0   & 0   & 200 \\
2 & \textsc{Balanced}    & 150         & 75  & 75  & 200 \\
3 & \textsc{Synth-Heavy} & 100         & 100 & 100 & 200 \\
\bottomrule
\end{tabular}
\end{table}

\paragraph{Reward design.}
A central challenge in training web agents on structured multi-field tasks is reward sparsity. Exact-match rewards are simple but assign zero credit unless the entire structured answer is correct. This is especially harsh for parallel web tasks, where an agent may retrieve several correct fields while missing others. We therefore study denser reward styles that provide partial credit for structured outputs, motivated by prior work on RLHF, process supervision, and GRPO while remaining mindful of specification-gaming risks \citep{ouyang2022training,lightman2024let,shao2024deepseekmath,amodei2016concrete}. In our reward-signal analysis, we compare exact-match scoring against an LLM-judge reward that assigns continuous correctness scores for partially correct answers. In our GRPO training runs, we use a structured partial-credit reward that combines answer correctness, format validity, and productive tool use. Answer correctness is computed recursively over structured outputs: dictionaries are scored over keys, lists are scored by matching each gold element against the best predicted element, and scalar values are scored using normalized exact match, containment, and token-level F1. This reward gives the policy denser feedback than an exact match while preserving a cheap non-LLM training loop for the main runs.

\paragraph{Agent scaffolds and baselines.}
We evaluate all systems on the same 150-example held-out set. Baselines use two scaffold families: a \textbf{WebWalker-style} scaffold with a single \texttt{visit\_page} tool, and a \textbf{WebExplorer-style} scaffold with \texttt{search} and \texttt{browse} tools using Jina Reader-style page extraction. These scaffolds follow the broader practice of exposing language models to browser or tool actions through a reasoning-action loop \citep{yao2023react,schick2023toolformer,nakano2021webgpt,zheng2024gpt,he2024webvoyager}. Under both scaffolds, we evaluate Qwen2.5-7B-Instruct, WebExplorer-8B, and GPT-4.1-mini. Decoding is greedy with \texttt{temperature}=0, \texttt{top\_p}=1, and \texttt{max\_new\_tokens}=4096.

\paragraph{Inference budgets.}
We evaluate each model under three inference budgets: 16k input context with 8 interaction rounds, 16k input context with 16 interaction rounds, and 64k input context with 16 interaction rounds. These settings let us separate three possible causes of failure: insufficient interaction rounds, insufficient context length, and failures that persist even with more budget.

\paragraph{Evaluation metrics.}
We report three main metrics. \textbf{Completion} measures whether the agent emits a parseable final answer rather than abstaining or timing out. \textbf{Binary Accuracy} measures whether the predicted answer fully satisfies the gold answer with no missing or contradictory information. \textbf{Element-wise F1} decomposes the gold and predicted answers into factual elements, matches semantically equivalent elements, and computes precision, recall, and F1. Following prior LLM-as-judge evaluation work and its documented caveats \citep{zheng2023judging,liu2023geval,wang2024large,kim2024prometheus}, Binary Accuracy and Element-wise F1 are judged by GPT-4.1-mini using the same evaluation protocol for all models and baselines. We also report not-attempted rate, content-filter/judge-failure rate, round usage, and precision-recall breakdowns as diagnostic metrics.

\paragraph{Reward signal analysis.}
We also compare exact-match and partial-credit reward signals to motivate the training objective. Exact match is sparse for structured multi-field answers because any missing or misformatted field can collapse the reward to zero. Partial-credit scoring provides denser feedback by assigning credit to correctly recovered answer elements. We report detailed reward-signal plots in Appendix~\ref{app:reward_signal}. The main results focus on whether these reward choices translate into reliable task completion and correctness.

\begin{table*}[t]
\centering
\small
\setlength{\tabcolsep}{4pt}
\caption{Baseline and GRPO-trained model results under three inference budgets. Column groups correspond to 16k input with $r{=}8$, 16k input with $r{=}16$, and 64k input with $r{=}16$, displaying completion rate, binary accuracy, and element-wise F1 (EW F1) scores. Completion is computed directly from model outputs. Binary Accuracy and Element-Wise F1 are judged by GPT-4.1-mini using the same evaluation protocol for all systems.}\label{tab:baseline_all}
\begin{tabularx}{\textwidth}{@{}Xccccccccc@{}}
\toprule
& \multicolumn{3}{c}{\textbf{16k input, $r{=}8$}} & \multicolumn{3}{c}{\textbf{16k input, $r{=}16$}} & \multicolumn{3}{c}{\textbf{64k input, $r{=}16$}} \\
\cmidrule(lr){2-4}\cmidrule(lr){5-7}\cmidrule(lr){8-10}
\textbf{Model} & \textbf{Comp.} & \textbf{Bin. Acc.} & \textbf{EW F1} & \textbf{Comp.} & \textbf{Bin. Acc.} & \textbf{EW F1} & \textbf{Comp.} & \textbf{Bin. Acc.} & \textbf{EW F1} \\
\midrule
\multicolumn{10}{@{}l}{\textit{WebWalker-style baselines}} \\
GPT-4.1-mini & 60.7 & 18.0 & 0.3229 & 65.3 & 19.3 & 0.3336 & 63.3 & 17.3 & 0.3220 \\
Qwen2.5-7B-Instruct & 29.3 & 4.7 & 0.1120 & 29.3 & 4.7 & 0.1209 & 27.3 & 5.3 & 0.1281 \\
WebExplorer-8B & 19.3 & 6.0 & 0.0922 & 24.0 & 4.7 & 0.0897 & 22.0 & 4.7 & 0.0998 \\
\midrule
\multicolumn{10}{@{}l}{\textit{WebExplorer-style baselines}} \\
GPT-4.1-mini & 100 & 27.3 & 0.4088 & 99.3 & 28.0 & 0.3949 & 100 & 27.3 & 0.2896 \\
Qwen2.5-7B-Instruct & 73.3 & 7.3 & 0.1861 & 72.0 & 8.0 & 0.2108 & 92.7 & 14.0 & 0.2675 \\
WebExplorer-8B & 37.3 & 16.0 & 0.1888 & 50.7 & 22.0 & 0.2489 & 55.3 & 20.7 & 0.2976 \\
\midrule
\multicolumn{10}{@{}l}{\textit{Our model (GRPO-trained)}} \\
Ours (\textsc{Human})          & 74.7 & 29.3 & 0.3832 & 88.7 & 31.3 & 0.4414 & 93.3 & 27.3 & 0.4288 \\
Ours (\textsc{Balanced})       & 74.7 & 36.0 & 0.3944 & 91.3 & 30.7 & 0.4368 & 91.3 & 30.7 & 0.4299 \\
Ours (\textsc{Synth-Heavy})    & \textbf{96.0} & 28.0 & 0.4038 & \textbf{96.0} & \textbf{33.3} & \textbf{0.4529} & \textbf{99.3} & 32.0 & 0.4313 \\
\bottomrule
\end{tabularx}
\end{table*}

Table~\ref{tab:baseline_all} shows that GRPO training substantially improves both completion and partial correctness over the base WebExplorer-8B model. Under the WebExplorer-style scaffold at 16k context and 16 interaction rounds, the base model completes 50.7\% of tasks, achieves 22.0\% binary accuracy, and reaches 0.2489 element-wise F1. The best GRPO model, \textsc{Synth-Heavy}, completes 96.0\% of tasks, reaches 33.3\% binary accuracy, and obtains 0.4529 element-wise F1. Thus, synthetic-heavy GRPO nearly doubles completion and substantially improves element-wise correctness.

However, the results also reveal a central failure boundary: improving completion is much easier than improving correctness. Across GRPO models, increasing inference budget often raises completion without producing a comparable gain in binary accuracy. For example, \textsc{Synth-Heavy} improves from 96.0\% completion at 16k/$r{=}16$ to 99.3\% completion at 64k/$r{=}16$, but its element-wise F1 decreases from 0.4529 to 0.4313. Similarly, \textsc{Human} reaches 93.3\% completion at 64k/$r{=}16$, but binary accuracy drops relative to its 16k/$r{=}16$ setting. These patterns suggest that longer context and larger interaction budgets are not reliable fixes for long-horizon web-agent errors.

The synthetic data mixture improves completion most clearly. \textsc{Synth-Heavy} has the highest completion in all three budget settings, reaching 96.0\%, 96.0\%, and 99.3\%. This suggests that synthetic supervision teaches the agent to continue interacting with the scaffold and emit final answers more reliably. Yet the gains in correctness are more limited. Binary accuracy remains near 30\%, and the best element-wise F1 is 0.4529. Synthetic data, therefore, reduces abstention and improves partial recovery, but it does not make the agent reliably correct.

The human-only and synthetic-only breakouts sharpen this conclusion. Models score higher on the held-out synthetic set than on the held-out human-verified set, confirming that synthetic examples are easier on average. This does not make the synthetic evaluation useless; instead, it provides a diagnostic for evaluation difficulty. Improvements that appear strong on the combined set must be interpreted alongside the human-only subset. On the human subset, \textsc{Synth-Heavy} achieves strong element-wise F1 but still has low binary accuracy, indicating that it often retrieves or emits some correct fields while failing to solve the full structured task.

Overall, GRPO with synthetic data is an effective intervention for reducing abstention and improving partial correctness. Still, it does not eliminate the main failure boundary: agents frequently produce complete-looking answers that are only partially supported by the trace.

\subsection{Failure Mode Analysis}
\label{sec:failure_modes}

Following the workshop's focus on reproducible triggers, trace diagnostics, and verified fixes, we analyze failures at the trajectory level rather than only through final scores. We focus on three recurring failure modes: context-bound search loops, premature termination on partial answers, and synthesis collapse. Each failure mode has an operational definition, a reproducible trigger, trace evidence, and a discussion of which interventions help or fail.

\paragraph{Aggregate evidence for the completion-correctness gap.}
Across all training mixes, scaling inference budget produces large gains in task completion but limited gains in answer correctness (Table~\ref{tab:baseline_all}). On the combined 150-example set, \textsc{Human} completion rises from 74.7\% at 16k/$r{=}8$ to 93.3\% at 64k/$r{=}16$, but element-wise F1 rises only from 0.3832 to 0.4288, and binary accuracy is non-monotone, dropping from 29.3\% to 27.3\%. \textsc{Synth-Heavy} achieves the highest completion, reaching 96.0\% to 99.3\% across budgets, but it does not consistently dominate on correctness. In particular, its element-wise F1 decreases from 0.4529 at 16k/$r{=}16$ to 0.4313 at 64k/$r{=}16$. This pattern shows that the main problem is not simply a lack of interaction budget. The trained agents often finish, but they do not reliably finish correctly.

Two additional diagnostics motivate our failure taxonomy. First, all three trained models are recall-heavy in element-wise scoring: precision is approximately 0.41-0.44, while recall is approximately 0.51-0.53. This indicates that agents often include too many elements in their final answers, some of which are unsupported. Second, round-usage distributions show that trained models usually terminate well below the 16-round budget, with median usage around 7-8 rounds and only 6-12 of 150 episodes reaching the cap. Thus, many failures are caused by premature termination or faulty synthesis rather than by running out of rounds. The three failure modes below decompose this completion-correctness gap.

\begin{table*}[t]
\centering
\small
\setlength{\tabcolsep}{4pt}
\caption{Failure-mode prevalence at 16k context, $r{=}16$, on the combined 150-example evaluation set. ``Imp.'' counts imperfect episodes with Binary Accuracy~$=0$. Failure-mode rates are computed over imperfect episodes rather than all 150 examples. Failure modes are not mutually exclusive, so ``Any'' counts episodes triggering at least one mode.}
\label{tab:fm_prevalence_main}
\begin{tabular}{@{}lcccccc@{}}
\toprule
\textbf{Model} & \textbf{N} & \textbf{Imp.} & \textbf{FM1 Loop} & \textbf{FM2 Early} & \textbf{FM3 Synth.} & \textbf{Any} \\
\midrule
Base SFT & 150 & 117 & 36 (31\%) & 16 (14\%) & 8 (7\%) & 54 (46\%) \\
Human & 150 & 103 & 42 (41\%) & 34 (33\%) & 20 (19\%) & 77 (75\%) \\
Balanced & 150 & 104 & 43 (41\%) & 29 (28\%) & 23 (22\%) & 82 (79\%) \\
Synth-Heavy & 150 & 100 & 25 (25\%) & 39 (39\%) & 18 (18\%) & 71 (71\%) \\
\bottomrule
\end{tabular}
\end{table*}

Table~\ref{tab:fm_prevalence_main} shows that the trace-level failure modes recur at scale rather than appearing only in isolated examples. GRPO-trained models have fewer imperfect episodes than the base model, but a larger fraction of their remaining errors fall into diagnosable failure modes. This is expected: after training, agents are more likely to emit final answers and produce analyzable traces, so errors shift from abstention and shallow failure toward search loops, premature partial answers, and synthesis failures. The \textsc{Synth-Heavy} run reduces FM1 relative to the other GRPO runs, but has the highest FM2 rate, supporting our central claim that synthetic-heavy GRPO improves answer emission more reliably than evidence-grounded completeness.

\begin{table*}[t]
\centering
\small
\setlength{\tabcolsep}{3pt}
\caption{Intervention effects on the completion-correctness gap. Aggregate results are on the 150-example combined evaluation set.}
\label{tab:intervention_summary}
\begin{tabularx}{\textwidth}{@{}p{2.2cm}p{3.8cm}p{3.0cm}X@{}}
\toprule
\textbf{Intervention} & \textbf{Aggregate effect} & \textbf{Failure-mode effect} & \textbf{Takeaway} \\
\midrule
GRPO training
& Base SFT $\rightarrow$ \textsc{Human}: Comp. +38.0, Bin. +9.3, element-wise F1 +0.1926
& Improves abstention and partial correctness; FM2/FM3 persist.
& Strongest overall intervention, but correctness remains far below completion. \\
\midrule
Synthetic-heavy data
& \textsc{Human} $\rightarrow$ \textsc{Synth-Heavy}: Comp. +7.3, Bin. +2.0, element-wise F1 +0.0115
& Mainly reduces abstention.
& Synthetic data teaches finishing more than verification. \\
\midrule
More rounds
& \textsc{Synth-Heavy} $r{=}8\rightarrow16$: Comp. +0.0, Bin. +5.3, element-wise F1 +0.0492
& Helps some partial-answer cases.
& Useful, but many failures terminate before the cap. \\
\midrule
64k context
& Across GRPO runs: completion weakly rises, but element-wise F1 drops by 0.0068-0.0216
& Partially helps Q4 FM1: element-wise F1 0.33$\rightarrow$0.67.
& Helps evidence retention, not synthesis or early stopping. \\
\midrule
Reward audit
& Q49 gets $R_{\text{total}}=0.8$ despite $R_{\text{correct}}=0$
& Explains FM2 reward-hacking pressure.
& Format/process credit should be gated on evidence-grounded correctness. \\
\bottomrule
\end{tabularx}
\end{table*}

\paragraph{FM1: Context-bound search loops.}
This mode is closest to retrieval-augmented and search-refreshed QA failures, where the system has access to external evidence. Still, it must retrieve, retain, and reuse the right snippets rather than repeatedly querying near-duplicates \citep{lewis2020retrieval,karpukhin2020dense,vu2024freshllms}.

\textit{Operational definition.}
A trajectory contains a contiguous run of $\geq 5$ search tool-calls
whose query strings share $\geq 3$ content words pairwise, retrieving
no new tool-response content not already present in earlier rounds, and
terminating either by a round-cap or by a final answer drawn from
pre-loop content.

\textit{Reproducible trigger.}
Multi-field aggregation questions where (a) gold attributes are not
literally named on canonical pages (e.g., a person's role expressed as
``founding editor'' on one page but queried as ``editorial board
member''), and (b) the relevant facts are retrieved in early rounds
under non-canonical phrasing. This trigger reproduces across training
mixes.

\textit{Trace evidence.}
Q4 (\emph{VLDB editorial roles 2010-2015}) reproduces FM1 across all
three runs: \textsc{Human} at 16k/$r{=}16$ uses 13 rounds, with
rounds 7-12 issuing variants of ``VLDB journal editorial board
2010-2015'' that returns the same DBLP page already retrieved in
round 3. \textsc{Balanced} at 16k/$r{=}16$ exhibits the same loop on
Q4 across 14 rounds, ultimately concluding that information was
``Not found on VLDB Journal editorial board'' for all six years
despite having retrieved Jagadish's PVLDB founding EiC role and
Weber's PVLDB Information Director role multiple times during the
trajectory. The agent searched for canonical-phrasing matches
(``VLDB Journal editorial board'') and discounted the retrieved
non-canonical evidence, confusing PVLDB and VLDB Journal as
distinct publications and producing a structurally complete but
factually empty answer. The Q4 element-wise F1 across budget conditions for
\textsc{Human} (0.33 at 16k, 0.67 at 64k) shows that larger context
mitigates severity by preserving early-round evidence, but does not
prevent the loop.

\textit{Budget and training-mix response.}
Larger context partially mitigates FM1 because earlier-retrieved
facts remain accessible at synthesis time, but the agent still
issues redundant searches. FM1 is not observed on the
synth-eval distribution,
indicating that the canonical phrasing
mismatch characteristic of human-curated questions drives the trigger. This is a distribution
property, not a training-mix property. \textsc{Synth-Heavy}
reproduces FM1 on human-eval despite training largely on synthetic
data.

\paragraph{FM2: Premature termination on partial answers.}
This mode connects to the tension between outcome-level credit, process-level credit, and reward hacking: dense rewards can reduce sparsity while still making incomplete but well-formed answers attractive \citep{lightman2024let,amodei2016concrete}.

\textit{Operational definition.}
The agent emits a special token (\texttt{<final\_answer>}) before exhausting at least
50\% of the round budget, with element-wise recall $< 0.6$ on the
gold answer. The trajectory's final-answer reward components
satisfy $R_{\text{format}} = 1$, $R_{\text{process}} \geq 0.9$,
yielding $R_{\text{total}} > 1.4$ despite incomplete coverage.

\textit{Reproducible trigger.}
List-valued questions (``list all $X$,'' ``every $Y$,'' ``for each
$Z$'') where the gold answer contains $\geq 4$ items spanning
heterogeneous source pages.

\textit{Trace evidence.}
Q47 (\emph{Angry Birds animated series 2022-2024}) reproduces FM2
across all three runs. \textsc{Human} at 16k/$r{=}16$ terminates in
5 rounds with 2/7 gold items retrieved (\emph{Summer Madness},
\emph{Mystery Island}); \textsc{Synth-Heavy} at 16k/$r{=}16$ also
terminates in 7 rounds with the same two items, missing
\emph{MakerSpace S3}, \emph{Slingshot Stories S3}, and others. Both
trajectories receive $R_{\text{total}} = 1.58$
($R_{\text{format}}=1.0$, $R_{\text{process}}=1.0$,
$R_{\text{correct}}=0.78$). Q49 (\emph{Illuvium reward
structure}) at \textsc{Synth-Heavy} 16k/$r{=}8$ terminates in 6
rounds with $R_{\text{correct}}=0$ but
$R_{\text{total}}=0.8$ — a non-final-answer trajectory would have
incurred a higher penalty under our reward function than the incorrect early termination did.

\textit{Budget and training-mix response.}
Premature termination is unaffected by budget scaling: the agents
terminate well before exhausting the round cap, regardless of
context size. FM2 reproduces on synth-eval (3 candidates across
runs), confirming it is a learned behavior rather than a
distribution artifact. \textsc{Synth-Heavy} has higher FM2 prevalence than \textsc{Human} at the main 16k/$r{=}16$ setting, consistent with the view that synthetic-heavy training improves answer emission but can increase early partial-answer failures.

\paragraph{FM3 — Synthesis collapse.}
This mode is a synthesis-side version of source attribution and atomic factuality failures: the evidence may be present, but the final answer no longer preserves the supported facts or their bindings \citep{gao2023enabling,rashkin2023measuring,min2023factscore}.

\textit{Operational definition.}
The agent's trajectory contains tool responses that are jointly sufficient
to produce an element-wise-correct answer, but the emitted
\texttt{<final\_answer>} either (i) replaces retrieved values with a single repeated default, (ii) replaces retrieved values
with ``Not found'' placeholders, or (iii) collapses
distinct-row schema fields to a homogeneous value. Element-wise F1 $< 0.6$
despite the trajectory having retrieved
$\geq 80\%$ of gold-relevant facts.

\textit{Reproducible trigger.}
Multi-row structured outputs (JSON arrays, per-year tables) where the schema admits a default value or where ``not found'' is
syntactically valid output.

\textit{Trace evidence.}
Q1 (\emph{ICSE 2002 submission deadlines}) at \textsc{Human}
64k/$r{=}16$ retrieves track-specific deadlines across rounds
1-9 (Technical Papers: Sep 25, 2001; Doctoral Symposium: Jan 6,
2002; Panels: Sep 18, 2001; Workshops/Tutorials/SOA: Oct 21,
2001), then collapses all eight tracks to ``October 21, 2001''
in the final answer. Element-wise F1 = 0.529 (partial credit for the three
Oct-21 tracks); Binary Accuracy = 0. Q42 (\emph{ETRA 2025
organizing committee}) at \textsc{Balanced} 16k/$r{=}16$
retrieves the full committee-roles list from the official
organization page, identifies ``Diversity and Accessibility
Chairs'' as the accessibility role, but emits
\texttt{total\_roles=19} when the gold count is 17. Q9 (\emph{DCMI 2024
schema representative}) at \textsc{Synth-Heavy} 16k/$r{=}16$
emits ``NOT FOUND'' placeholders for both fields after a
16-round trajectory, despite having retrieved relevant context.

\textit{Budget and training-mix response.}
FM3 is least responsive to budget scaling because the needed facts are already present, but the final answer fails to preserve them. FM3 reproduces on
synth-eval (2 candidates across runs) and on human-eval across
all three training mixes. The recall-heavy precision/recall
imbalance reported above (precision $\approx 0.43$, recall
$\approx 0.51$) is the population-scale signature of FM3:
agents over-include elements in answers without faithfully
preserving the structure of retrieved evidence.

\paragraph{Failure-mode prevalence by training mix.}
Aggregating across the 150-example combined eval at 16k/$r{=}16$,
the element-wise F1 distribution (Table~\ref{tab:ewf1_distribution})
decomposes into approximately consistent failure-mode shares
across runs. \textsc{Human}, \textsc{Balanced}, and
\textsc{Synth-Heavy} all produce 16-20\% Perfect (element-wise F1
$\geq 0.95$), 29-36\% High (0.5-0.95), 17-24\% Partial
(0-0.5), and 27-33\% Zero. The training mix shifts the
completion rate (Zero share drops from 33\% for
\textsc{Human} to 27\% for \textsc{Synth-Heavy}) but does not
substantially redistribute episodes between the failure modes.
This is direct evidence that the three failure modes are
properties of the training paradigm (GRPO with our reward
specification) rather than properties of any particular
training data source.

\paragraph{Repair implications.}
The main lesson is that answer emission and interaction length are weak proxies for web-agent reliability. Each failure mode points to a different repair target. FM1 requires query-diversity and evidence-retention checks, because a larger context can preserve early evidence but does not stop redundant search loops. FM2 requires coverage-gated stopping, since agents often terminate before exhausting the round budget with incomplete list-valued answers. FM3 requires evidence-bound synthesis, because the relevant facts may already appear in the trace but still be collapsed or misbound in the final answer. Thus, future fixes should combine partial-credit training with explicit coverage checks and final-answer verification rather than simply increasing context, rounds, or synthetic data, potentially drawing on ideas from decomposition, self-reflection, and process supervision~\cite {yao2023tree,shinn2023reflexion,lightman2024let}.

\section{Conclusion}
\label{sec:conclusion}

We presented \method, a benchmark and construction pipeline for parallelizable web exploration. The dataset combines 350 manually verified parallel tasks with 1,329 reconstructed records whose answers and URL-based trajectories are verified against the live web. This setting exposes a failure boundary that standard web-agent evaluation can miss: agents may complete the interaction and emit structured answers while still missing fields, over-including unsupported items, or collapsing retrieved evidence during final synthesis.

Our experiments show that GRPO and synthetic supervision improve completion and partial correctness, but trace diagnostics reveal that these gains do not eliminate core failure modes. More context and interaction budget improve completion without consistently improving binary accuracy, and synthetic-heavy training performs better on easier synthetic tasks than on harder human-verified tasks. Future web-agent training should therefore optimize not only for answer emission or broad retrieval, but also for evidence-grounded verification, aggregation constraints, and diagnostics that distinguish finishing from actually solving the task.

\section*{Impact Statement}
This paper studies failure modes in long-horizon web agents. The intended impact is to improve the reliability and diagnosability of agentic AI systems by identifying cases in which agents produce plausible final answers without fully supporting them with retrieved evidence. Better trace-level diagnostics may reduce the risk of deploying agents that silently omit or misinterpret information. The work also relies on automated judging and web-derived data, so results should be interpreted with attention to judge reliability, changing web content, and possible dataset biases.

\section*{Acknowledgments}

We gratefully acknowledge the support of the NSF Diamond project OAC-2311767 (Democratizing Large Neural Network Model Training for Science).

\bibliographystyle{icml2026}
\bibliography{references}

\clearpage
\appendix
\onecolumn
\raggedbottom

\section{LLM Judge Validation}
\label{app:judge_validation}

We validate the judge on a stratified subset of held-out predictions for Binary Accuracy and element-wise F1 produced by the LLM judge (GPT-4.1-mini). The audit samples predictions across model variants, correctness bands, and failure modes. Human annotators label binary correctness and element-level precision/recall using the same gold answers shown to the judge. The judge is blinded to the model identity and training condition. Most audited predictions are not fully correct, so the largest cell is the true-negative cell where both the human annotator and judge mark the answer incorrect.

\begin{table}[h!]
\centering
\small
\setlength{\tabcolsep}{5pt}
\caption{Manual validation of the GPT-4.1-mini judge on a stratified subset of held-out predictions. Human annotators label binary correctness and element-wise F1 using the same gold answers shown to the judge.}
\label{tab:judge_validation}
\begin{tabular}{@{}lcccccc@{}}
\toprule
\textbf{N} & \textbf{Human F1} & \textbf{Judge F1} & \textbf{MAE} & \textbf{Pearson} & \textbf{Spearman} & \textbf{Binary Agree} \\
\midrule
50 & 0.3751 & 0.3548 & 0.0751 & 0.8674 & 0.8774 & 96.0\% \\
\bottomrule
\end{tabular}
\end{table}

\begin{table}[h!]
\centering
\small
\setlength{\tabcolsep}{8pt}
\caption{Binary correctness confusion matrix for GPT-4.1-mini judge validation. Rows are judge labels and columns are human labels. The judge agrees with humans on 48 of 50 examples.}\label{tab:judge_confusion}
\begin{tabular}{@{}lcc@{}}
\toprule
 & \textbf{Human Correct} & \textbf{Human Incorrect} \\
\midrule
\textbf{Judge Correct} & 6 & 1 \\
\textbf{Judge Incorrect} & 1 & 42 \\
\bottomrule
\end{tabular}
\end{table}

The judge validation audit shows strong agreement between human annotation and GPT-4.1-mini on a stratified 50-example subset.
The judge's mean element-wise F1 score differs from the human mean by only 0.0203, with an MAE of 0.0751.
Rank correlation is high across examples, with Pearson's correlation of 0.8674 and Spearman's correlation of 0.8774.
For binary correctness, the judge matches human labels on 48 out of 50 examples, yielding 96.0\% agreement and Cohen's $\kappa=0.8339$.
The two disagreements consist of one false positive and one false negative.
This supports using the judge for aggregate comparisons. At the same time, the main conclusions remain based on trends in completion rates, element-wise score distributions, and trace-level failure analysis.

\section{Human and Synthetic Breakout Results}
\label{app:breakout_results}

Tables~\ref{tab:ewf1_distribution} and~\ref{tab:human_synth_breakout} provide the full breakout behind the aggregate results in Section~\ref{sec:failure_modes}. Table~\ref{tab:ewf1_distribution} shows that GRPO shifts many examples out of the zero-score bucket: the base WebExplorer-8B model has 87/150 zero-EW-F1 outputs, while \textsc{Synth-Heavy} has 40/150. However, many examples remain in the High or Partial bands rather than becoming fully correct, which supports the completion-correctness gap discussed in the main text. Table~\ref{tab:human_synth_breakout} shows that held-out synthetic examples are easier than human-verified examples. For \textsc{Synth-Heavy} at 16k/$r{=}16$, human examples reach 92.0\% completion, 14.0\% binary accuracy, and 0.4276 element-wise F1, while synthetic examples reach 98.0\% completion, 43.0\% binary accuracy, and 0.4655 element-wise F1.

\begin{table}[h!]\centering
\small
\setlength{\tabcolsep}{4pt}
\caption{Distribution of EW F1 (element-wise F1) scores across the 150-example combined evaluation at 16k context, $r{=}16$. Each model's outputs are bucketed by element-wise F1: Perfect ($\geq 0.95$), High ($[0.5, 0.95)$), Partial ($(0, 0.5)$), and Zero. Counts are followed by percentages of $N{=}150$.}
\label{tab:ewf1_distribution}
\begin{tabular}{@{}lcccc@{}}
\toprule
\textbf{Model} & \textbf{Perfect} & \textbf{High} & \textbf{Partial} & \textbf{Zero} \\
 & ($\geq 0.95$) & ($[0.5, 0.95)$) & ($(0, 0.5)$) & ($=0$) \\
\midrule
Base SFT / WebExplorer-8B & 16/150 (10.7\%) & 22/150 (14.7\%) & 25/150 (16.7\%) & 87/150 (58.0\%) \\
Ours (Split 1: Human-only) & 27/150 (18.0\%) & 48/150 (32.0\%) & 25/150 (16.7\%) & 50/150 (33.3\%) \\
Ours (Split 2: Balanced) & 23/150 (15.3\%) & 54/150 (36.0\%) & 29/150 (19.3\%) & 44/150 (29.3\%) \\
Ours (Split 3: Synth-Heavy) & 30/150 (20.0\%) & 44/150 (29.3\%) & 36/150 (24.0\%) & 40/150 (26.7\%) \\
\bottomrule
\end{tabular}
\end{table}

\begin{table}[h!]\centering
\setlength{\tabcolsep}{4pt}
\caption{Performance breakout by evaluation subset (16k context, $r{=}16$). The held-out human-verified subset (50 examples) and the held-out synthetic subset (100 examples) are reported separately for each model. Completion measures answer emission; Binary Accuracy and element-wise F1 are GPT-4.1-mini-judged.}
\label{tab:human_synth_breakout}
\begin{tabular}{@{}llrccc@{}}
\toprule
\textbf{Model} & \textbf{Subset} & \textbf{N} & \textbf{Comp.} & \textbf{Bin. Acc.} & \textbf{element-wise F1} \\
\midrule
Base SFT / WebExplorer-8B & Human & 50 & 38.0\% & 6.0\% & 0.1717 \\
 & Synth & 100 & 57.0\% & 30.0\% & 0.2875 \\
Ours (Split 1: Human-only) & Human & 50 & 76.0\% & 10.0\% & 0.3622 \\
 & Synth & 100 & 95.0\% & 42.0\% & 0.4811 \\
Ours (Split 2: Balanced) & Human & 50 & 78.0\% & 16.0\% & 0.3659 \\
 & Synth & 100 & 98.0\% & 38.0\% & 0.4722 \\
Ours (Split 3: Synth-Heavy) & Human & 50 & 92.0\% & 14.0\% & 0.4276 \\
 & Synth & 100 & 98.0\% & 43.0\% & 0.4655 \\
\bottomrule
\end{tabular}
\end{table}

\FloatBarrier
\section{Training Hyperparameters}
\label{app:training_hparams}

Table~\ref{tab:grpo_hparams} reports the shared training configuration for the three GRPO runs in Table~\ref{tab:data_splits}. All three runs use the same base model, scaffold, rollout budget, optimizer settings, and reward weights; only the training data mixture changes.

\begin{table}[h!]
\centering
\small
\setlength{\tabcolsep}{5pt}
\caption{GRPO training hyperparameters. All trained runs use the same configuration; only the human/synthetic data mixture differs.}
\label{tab:grpo_hparams}
\begin{tabular}{@{}ll@{}}
\toprule
\textbf{Parameter} & \textbf{Value} \\
\midrule
Base model & WebExplorer-8B \\
Training algorithm & GRPO-style clipped policy optimization \\
Trainable parameters & LoRA adapters \\
LoRA rank / alpha & 64 / 128 \\
Training steps & 200 \\
Generations per prompt & 6 \\
Rollout scaffold & WebExplorer-style search/browse scaffold \\
Rollout context length & 16k tokens \\
Rollout interaction budget & 8 rounds \\
Maximum new tokens & 4096 \\
Sampling temperature / top-$p$ & 0.8 / 0.95 \\
Learning rate & $1\times 10^{-5}$ \\
Optimizer & AdamW \\
Weight decay & 0.0 \\
KL coefficient $\beta$ & 0.01 \\
Clip range $\epsilon$ & 0.2 \\
Entropy bonus & 0.01 \\
Gradient clipping & 1.0 \\
Warmup ratio & 0.03 \\
Precision & bfloat16 \\
Reward & $R_{\text{correct}} + 0.3R_{\text{format}} + 0.5R_{\text{process}}$ \\
Correctness reward & Recursive dictionary/list/scalar fuzzy matching \\
Format reward & Structural tag validity \\
Process reward & Productive tool-use heuristic \\
Checkpoint frequency & Every 5 updates \\
\bottomrule
\end{tabular}
\end{table}

Our scaffold follows the search/browse ReAct interface used by recent long-horizon web-agent systems, while our experiment differs by holding the scaffold fixed and varying the human/synthetic training mixture to study failure modes rather than maximizing benchmark accuracy.

\section{Benchmark Construction Details}
\label{app:construction_details}

\paragraph{Manual annotation.}
Human annotators opened each root URL, checked whether the requested information was reachable from the root website, and recorded the sequential step count, maximum parallel degree, visited URLs grouped by step, and the golden answer in JSON format. Partially answerable questions were repaired when possible; fully unanswerable questions were marked with a short failure description.

\paragraph{Unanswerable question categories.}
During manual verification, we observed six recurring construction-time failure categories: stale or removed content, incorrect root URL, missing fine-grained details, cross-site information conflation, subjective or non-deterministic answers, and excessively long answers. These categories are used for dataset filtering and are distinct from the model failure modes analyzed in Section~\ref{sec:failure_modes}.

\paragraph{Trace generation.}
After verification, question-answer pairs are converted into ReAct-style trajectories using a web-exploration model prompted to produce reasoning traces consistent with the human-validated answer. Because the endpoint answer and website evidence are already verified, this stage produces structured supervision rather than inferring correctness from scratch.

\paragraph{Reconstruction details.}
For Category~B records, path reconstruction operates without using the original text labels. A GPT-4.1-based agent searches from the root URL to each target source page, follows listing pages and pagination when needed, and falls back to keyword search when navigation is unproductive. Per-source paths are merged via depth-first alignment, so shared intermediate nodes appear only once, while distinct targets appear in parallel. Answers are re-extracted from verified source-page content rather than inherited from the original dataset.

\section{Path Verification Details}
\label{app:path_verification}

Every candidate navigation path produced by the reconstruction pipeline undergoes strict link-chain verification. For each consecutive URL pair $(u_i, u_{i+1})$ in a proposed path, the system extracts all hyperlinks from the page at $u_i$ and confirms that $u_{i+1}$ appears among them. If verification fails for any pair, the system attempts to repair the path using breadth-first search over the cached link graph. Starting from $u_i$, BFS explores linked pages up to a depth of three, searching for an alternative route to $u_{i+1}$. If BFS succeeds, the repaired sub-path replaces the broken segment. If BFS fails, the entire path is marked as unverifiable, and the corresponding record is excluded.

This strict verification ensures that every reconstructed trajectory corresponds to a real, navigable sequence of hyperlinks on the live web at the time of construction.

\section{Reward Signal Analysis}
\label{app:reward_signal}

A central challenge in training web agents on structured multi-field tasks is reward sparsity \citep{ouyang2022training,lightman2024let,amodei2016concrete}.
Exact-match rewards are simple and strict, but they assign zero credit unless the entire structured output is correct.
This is especially harsh for parallel web-exploration tasks, where an agent may recover several correct fields while missing or misformatting others.
We therefore use this appendix to compare exact-match scoring with denser partial-credit scoring.
This diagnostic should not be confused with our final evaluation protocol or our main GRPO reward. 
GPT-4.1-mini is used only for the final held-out evaluation, where it scores all models under the same blinded protocol. 
The reward-signal plot here uses WebExplorer-8B as an open-model judge for a diagnostic comparison against an exact match. 
The final three GRPO runs in the main experiments use a structured fuzzy partial-credit reward rather than GPT-4.1-mini reward feedback.

\begin{figure}[H]
    \centering
    \includegraphics[width=0.86\textwidth]{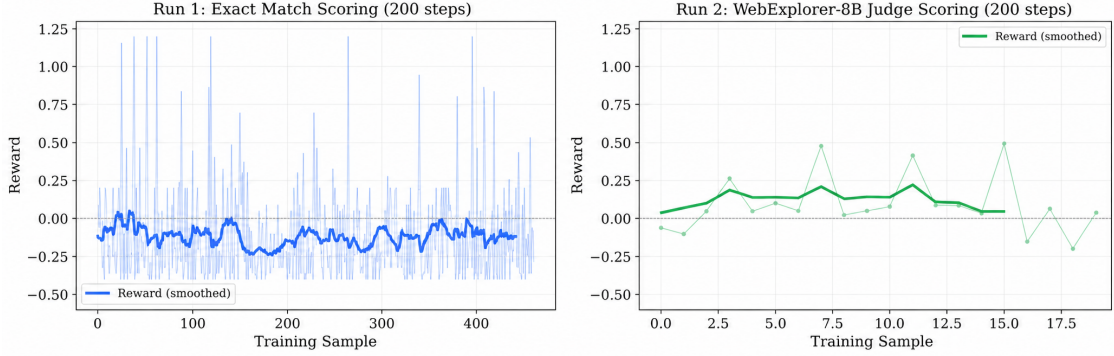}
    \caption{Reward signal comparison across training samples. \textbf{Left}: Exact-match scoring yields sparse rewards with infrequent positive spikes. \textbf{Right}: WebExplorer-8B judge scoring produces a denser and smoother reward signal. The x-axis represents individual training samples, and the y-axis represents the reward value.}
    \label{fig:reward_signal}
\end{figure}

\begin{figure}[H]
    \centering
    \includegraphics[width=0.86\textwidth]{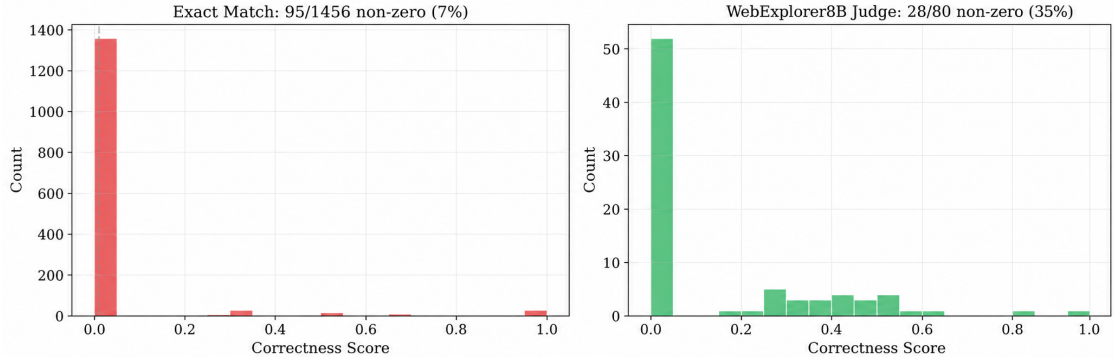}
    \caption{Correctness score distribution by reward function. Exact-match scoring produces non-zero correctness scores for 95 out of 1,456 evaluated samples (6.5\%), while the WebExplorer-8B judge scoring sample produces non-zero scores for 28 out of 80 samples (35.0\%). Because the sample sizes differ, we use this as a diagnostic of reward sparsity rather than a controlled ablation.}
    \label{fig:correctness_distribution}
\end{figure}

Figure~\ref{fig:reward_signal} compares the reward traces produced by two scoring styles: exact-match scoring and WebExplorer-8B judge scoring.
Exact-match scoring produces a sparse signal with infrequent positive events, while the WebExplorer-8B judge sample produces more graded non-zero scores.
Figure~\ref{fig:correctness_distribution} shows the same pattern in distributional form: exact-match scoring yields non-zero scores for 95 out of 1,456 evaluated samples (6.5\%), while the WebExplorer-8B judge scoring sample yields non-zero scores for 28 out of 80 samples (35.0\%).
Because these samples differ in size, we treat this comparison as a reward-sparsity diagnostic rather than a controlled training ablation.
The result supports the motivation for partial-credit rewards, but should not be read as isolating the effect of the judge itself.

The practical takeaway is that an exact match alone is a poor training signal for multi-field web tasks.
If an answer contains five fields and four are correct, an exact match still gives zero reward.
A partial-credit reward can assign credit to the recovered fields and provide the policy optimizer with a more informative learning signal.
This motivates the structured reward used in our GRPO runs, which scores dictionaries, lists, and scalar fields recursively rather than relying only on an all-or-nothing exact match.

This reward analysis also connects to the failure modes in the main paper.
Dense rewards help reduce abstention and encourage final-answer emission, but they can also reward incomplete answers if format and process terms are not gated by evidence-grounded correctness \citep{amodei2016concrete,lightman2024let}.
This is visible in FM2, where agents sometimes terminate early with a well-formed but incomplete answer.
Thus, partial-credit rewards are useful, but they should be paired with coverage checks and evidence-grounded verification rather than treated as a complete fix.

\FloatBarrier

\section{Full Intervention Evidence}
\label{app:full_intervention}

\begin{table}[t]
\centering
\small
\setlength{\tabcolsep}{4pt}
\caption{Failure-mode prevalence across inference budgets on the combined 150-example evaluation set. ``Imp.'' counts imperfect episodes with Binary Accuracy~$=0$. Failure-mode rates are computed over imperfect episodes. Failure modes are not mutually exclusive.}
\label{tab:fm_prevalence_appendix}
\begin{tabular}{@{}llcccccc@{}}
\toprule
\textbf{Model} & \textbf{Budget} & \textbf{N} & \textbf{Imp.} & \textbf{FM1 Loop} & \textbf{FM2 Early} & \textbf{FM3 Synth.} & \textbf{Any} \\
\midrule
Base SFT & 16k, $r=8$ & 150 & 126 & 33 (26\%) & 5 (4\%) & 6 (5\%) & 44 (35\%) \\
Base SFT & 16k, $r=16$ & 150 & 117 & 36 (31\%) & 16 (14\%) & 8 (7\%) & 54 (46\%) \\
Base SFT & 64k, $r=16$ & 150 & 119 & 35 (29\%) & 20 (17\%) & 13 (11\%) & 58 (49\%) \\
\midrule
Human & 16k, $r=8$ & 150 & 106 & 38 (36\%) & 8 (8\%) & 14 (13\%) & 57 (54\%) \\
Human & 16k, $r=16$ & 150 & 103 & 42 (41\%) & 34 (33\%) & 20 (19\%) & 77 (75\%) \\
Human & 64k, $r=16$ & 150 & 109 & 45 (41\%) & 42 (39\%) & 21 (19\%) & 89 (82\%) \\
\midrule
Balanced & 16k, $r=8$ & 150 & 96 & 39 (41\%) & 7 (7\%) & 9 (9\%) & 51 (53\%) \\
Balanced & 16k, $r=16$ & 150 & 104 & 43 (41\%) & 29 (28\%) & 23 (22\%) & 82 (79\%) \\
Balanced & 64k, $r=16$ & 150 & 104 & 42 (40\%) & 31 (30\%) & 17 (16\%) & 79 (76\%) \\
\midrule
Synth-Heavy & 16k, $r=8$ & 150 & 108 & 21 (19\%) & 11 (10\%) & 13 (12\%) & 42 (39\%) \\
Synth-Heavy & 16k, $r=16$ & 150 & 100 & 25 (25\%) & 39 (39\%) & 18 (18\%) & 71 (71\%) \\
Synth-Heavy & 64k, $r=16$ & 150 & 102 & 22 (22\%) & 50 (49\%) & 15 (15\%) & 74 (73\%) \\
\bottomrule
\end{tabular}
\end{table}

\paragraph{Mitigations and trade-offs.}
Each failure mode admits distinct interventions, with
asymmetric trade-offs. (1)~\textit{Reward reshaping}: the
composite reward
$R = 0.3\cdot R_{\text{format}} + R_{\text{correct}} + 0.5
\cdot R_{\text{process}}$ permits trajectories to receive
$R \geq 0.8$ with $R_{\text{correct}} = 0$
(Q49 trajectory above). Reducing $w_{\text{format}}$ and
$w_{\text{process}}$ to $\leq 0.1$ each, and gating
$R_{\text{process}}$ on non-zero correctness would shrink the
reward-hacking surface that drives FM2 specifically. The
trade-off is reward sparsity, which Appendix~\ref{app:reward_signal} showed is itself problematic. (2)~\textit{Context scaling}
mitigates FM1 (Q4 element-wise F1: 0.33$\to$0.67 from 16k to 64k) but does not eliminate the loop and does not affect FM2 or FM3. The trade-off is compute, and the marginal recall gain saturates by
$r{=}16$ (median round usage is 7-8 across all conditions).
(3)~\textit{Parallel decomposition} addresses FM1 and FM2
structurally: per-year sub-agents on Q4 eliminate cross-year
search interference, and per-item sub-agents on Q47 force
completeness checks at the orchestrator level. FM3 is not
addressed by parallel decomposition alone; it requires either a verification step over the retrieved content at synthesis
time or structured generation constraints that bind emitted
fields to retrieved evidence. (4)~\textit{Synthetic data},
contrary to one possible reading of our results, does not
introduce new failure modes — \textsc{Synth-Heavy} reproduces
all three modes — but improves coverage on completion. This
suggests that synthetic data and reward reshaping are
complementary rather than substitutable interventions.

\begin{table}[t]
\centering
\small
\setlength{\tabcolsep}{5pt}
\caption{Intervention effects on the completion-correctness gap. Checkmarks indicate clear aggregate improvement, triangles indicate partial or failure-mode-specific improvement, and crosses indicate no consistent improvement.}
\label{tab:intervention_by_failure}
\begin{tabular}{@{}p{3.0cm}p{9.7cm}c@{}}
\toprule
\textbf{Intervention} & \textbf{Evidence and interpretation} & \textbf{Verdict} \\
\midrule
GRPO training
& Base SFT $\rightarrow$ \textsc{Human} at 16k/$r{=}16$: completion improves 50.7$\rightarrow$88.7 (+38.0), binary accuracy improves 22.0$\rightarrow$31.3 (+9.3), and element-wise F1 improves 0.2489$\rightarrow$0.4414 (+0.1926). GRPO is the strongest overall intervention, but FM2/FM3 persists, and binary correctness remains far below the completion rate.
& \checkmark \\
\midrule
Synthetic-heavy data
& \textsc{Human} $\rightarrow$ \textsc{Synth-Heavy} at 16k/$r{=}16$: completion improves 88.7$\rightarrow$96.0 (+7.3), binary accuracy improves 31.3$\rightarrow$33.3 (+2.0), and element-wise F1 improves 0.4414$\rightarrow$0.4529 (+0.0115). Synthetic-heavy training mainly teaches the model to finish more reliably; correctness gains are smaller and stronger on synthetic examples.
& $\triangle$ \\
\midrule
More rounds
& \textsc{Synth-Heavy} from 16k/$r{=}8$ to 16k/$r{=}16$: completion stays 96.0, binary accuracy improves 28.0$\rightarrow$33.3 (+5.3), and element-wise F1 improves 0.4038$\rightarrow$0.4529 (+0.0492). More rounds help some partial-answer cases, but many failures stop before the round cap.
& $\triangle$ \\
\midrule
64k context
& From 16k/$r{=}16$ to 64k/$r{=}16$, completion weakly rises across GRPO runs, but element-wise F1 drops by 0.0068-0.0216. Larger context partly mitigates FM1 in Q4, where element-wise F1 rises from 0.33 to 0.67, but it does not improve aggregate correctness.
& $\triangle$ \\
\midrule
Reward audit
& Q49 receives $R_{\text{total}}=0.8$ despite $R_{\text{correct}}=0$. Format and process rewards can make an incorrect early-final trajectory less costly than abstention, suggesting that future reward designs should gate process credit to evidence-grounded answer correctness.
& $\times$ \\
\bottomrule
\end{tabular}
\end{table}

Table~\ref{tab:intervention_by_failure} shows that the interventions improve different parts of the failure boundary. GRPO training produces the largest aggregate gain, improving completion by 38.0 points and element-wise F1 by 0.1926 over the base WebExplorer-8B model at 16k/$r{=}16$. Synthetic-heavy data further improves completion, but its gains in correctness are small. Increasing the round budget improves \textsc{Synth-Heavy} correctness, but median round usage remains only 7-8 of 16, so many failures are not due to exhausting the interaction budget. Increasing context to 64k partially mitigates specific context-bound search loops, such as Q4, but does not improve aggregate element-wise F1. Finally, the Q49 reward audit shows why premature termination can persist: a trajectory can receive substantial format and process credit even when its final answer is incorrect.

\section{Annotation Guideline and Prompts}
\label{app:annotation_prompts}

The annotation guideline provided to annotators followed this workflow:
\begin{enumerate}[nosep]
    \item Open the root URL in a web browser.
    \item Read the generated question and determine whether the requested information is available from pages reachable from the root URL.
    \item If answerable, navigate to the relevant subpages and record the sequential step count, maximum parallel degree, visited URLs grouped by step, and golden answer in JSON format.
    \item If partially answerable, note the missing information and propose a minimally modified version of the question that becomes fully answerable.
    \item If unanswerable, mark the question as such and briefly describe the failure mode using the construction-time categories in Appendix~\ref{app:construction_details}.
\end{enumerate}
This protocol follows the same evidence-first approach as fact-verification and attribution-annotation frameworks \citep{thorne2018fever,rashkin2023measuring}.

Detailed prompts for question generation, agentic verification, path reconstruction, and judge evaluation are included in the supplementary materials.

\section{Example Questions Used in Trace Diagnostics}
\label{app:example_questions}

Below are the full prompts for the question IDs referenced in the trace-diagnostic case studies. Complete trajectories, tool observations, and structured gold answers are included in the supplementary materials.

\begin{enumerate}[leftmargin=*]
    \item \textbf{Q4, FM1, VLDB editorial roles.}
    For each year from 2010 to 2015, list the editorial roles held by H. V. Jagadish and Gerald Weber in the VLDB journal.
    \textbf{Return a JSON array of objects with fields \texttt{\{year, H\_V\_Jagadish\_role, Gerald\_Weber\_role\}}.}

    \item \textbf{Q47, FM2, Angry Birds animated series.}
    What is the official theme of RovioCon Google 2024, and what is the title of every Angry Birds animated series released between 2022 and 2024?
    \textbf{Return a JSON object with fields \texttt{\{rovio\_con\_2024\_theme, animated\_series\_titles\}}.}

    \item \textbf{Q49, FM2 / reward audit, Illuvium rewards.}
    Compare the reward structure for Illuvium Beyond before and after the adoption of IIP-67 and IIP-69.
    \textbf{Return a table with columns: \texttt{\{period (before/after), reward\_structure\_description\}}.}

    \item \textbf{Q1, FM3, ICSE 2002 deadlines.}
    List all submission deadlines for every track, for example technical papers, doctoral symposium, workshops, and tutorials, at ICSE 2002.
    \textbf{Return a JSON array of objects with fields \texttt{\{track, deadline\}}.}

    \item \textbf{Q42, FM3, ETRA 2025 organizing committee.}
    How many distinct organizing committee roles are there for ETRA 2025, and which role is responsible for accessibility?
    \textbf{Return a JSON object with fields \texttt{\{total\_roles, accessibility\_role\}}.}

    \item \textbf{Q9, FM3, DCMI 2024 schema.}
    During the DCMI 2024 event, what is the promotion code for booking discounted flights with Air Canada, and which prominent schema.org representative is associated with this event's planning and outreach activities?
\end{enumerate}

\end{document}